\documentclass[runningheads]{llncs}
\usepackage[utf8]{inputenc}
\usepackage[misc]{ifsym}
\usepackage{paralist,soul,framed,enumitem,todonotes,xcolor,verbatim}
\usepackage{graphicx,subcaption}
\usepackage[colorlinks=true, allcolors=blue]{hyperref}
\usepackage{caption}
\usepackage{rotating}
\usepackage{wrapfig}
\usepackage{authblk}
\usepackage{booktabs}

\title{Explaining Deep Learning Hidden Neuron Activations using Concept Induction}

\author{Abhilekha Dalal\inst{1} \and Md Kamruzzaman Sarker\inst{2} \and Adrita Barua\inst{1} \and Pascal Hitzler\inst{1}}

\institute{Department of Computer Science, Kansas State University, USA \and
Department of Computing Sciences, University of Hartford, USA
\email{\ adalal@ksu.edu, sarker@hartford.edu, adrita@ksu.edu, hitzler@ksu.edu}}

\authorrunning{Dalal, A., Sarker, M., Barua, A, and Hitzler, P.}

\titlerunning{Explaining Hidden Neuron Activations using Concept Induction}

\begin{document}

\maketitle

\begin{abstract}

One of the current key challenges in Explainable AI is in correctly interpreting activations of hidden neurons. It seems evident that accurate interpretations thereof would provide insights into the question what a deep learning system has internally \emph{detected} as relevant on the input, thus lifting some of the black box character of deep learning systems.

The state of the art on this front indicates that hidden node activations appear to be interpretable in a way that makes sense to humans, at least in some cases. Yet, systematic automated methods that would be able to first hypothesize an interpretation of hidden neuron activations, and then verify it, are mostly missing. 

In this paper, we provide such a method and demonstrate that it provides meaningful interpretations. It is based on using large-scale background knowledge -- a class hierarchy of approx. 2 million classes curated from the Wikipedia Concept Hierarchy -- together with a symbolic reasoning approach called \emph{concept induction} based on description logics that was originally developed for applications in the Semantic Web field. 

Our results show that we can automatically attach meaningful labels from the background knowledge to individual neurons in the dense layer of a Convolutional Neural Network through a hypothesis and verification process. 
\end{abstract}

\section{Introduction}
\label{introduction}
The origins of Artificial Intelligence trace back several decades ago, and AI has been successfully applied to multiple complex tasks such as image classification \cite {ramprasath2018image}, speech recognition \cite{graves2014towards}, language translation \cite{auli2013joint}, drug design \cite{segler2018generating}, treatment diagnosis \cite{choi2019artificial}, and climate sciences \cite{liu2016application}, as an instance for just a few. Artificially intelligent machines reach exceptional performance levels in learning to solve more and more complex computational problems by possessing the capabilities of learning, thinking, and adapting -- mimicking human behavior to some extent, making them crucial for future development. 

Despite their success in a wide variety of tasks, there is a general distrust of their results. Powerful AI machines particularly Deep Neural Networks, are hard to explain and are often referred to as "Black Boxes" simply because there are no clear human-understandable explanations as to why the network gave the particular output. Many cases have been reported; for example, In 2019 Apple co-founder Steve Wozniak accused Apple Card of gender discrimination, claiming that the card gave him a credit limit that was ten times higher than that of his wife, even though the couple shares all property.\footnote{https://worldline.com/en/home/knowledgehub/blog/2021/january/ever-heard-of-the-aI-black-box-problem.html}. In CEO image search, while 27\% of US CEOs were women, only 11\% of the top image results for “CEOs” were featured as women.\footnote{https://www.mckinsey.com/featured-insights/artificial-intelligence/tackling-bias-in-artificial-intelligence-and-in-humans} In continuation to the mentioned observation, the output of a network's classification can be altered by introducing Adversarial examples \cite{bau2020understanding}, and there are many more attack techniques. It becomes a need to understand the reasoning behind how a system behaves and generates an output in a human-interpretable way, especially since the popularity of these systems has grown to such an extent that these systems are responsible for decisions previously taken by human beings in safety-critical situations, for example like self-driving cars \cite{chen2017end}, drug discovery and treatment recommendations \cite{rifaioglu2020deepscreen,hariri2021deep}. 

Explainable AI has been pursued for several years already, and the quest for efficient algorithms to generate human-understandable explanations has led to a significant number of contributions based on different approaches. 
or internal unit summarizing \cite{zhou2018interpreting,bau2020understanding}. Improvements in deep learning show that neurons in the hidden layer of the neural network can detect human-interpretable concepts that were not explicitly taught to the network, such as objects, parts, gender, context, sentiment etc \cite{bau2018identifying,karpathy2015visualizing,qi2017pointnet}.

In our approach which we present in this paper, we make central use of \emph{concept induction} \cite{DBLP:journals/ml/LehmannH10}, which has been developed for use in the Semantic Web field and is based on deductive reasoning over description logics, i.e., over logics relevant to ontologies, knowledge graphs and generally the Semantic Web field \cite{DBLP:books/crc/Hitzler2010,DBLP:journals/cacm/Hitzler21}. In a nutshell -- and more details are given below -- a concept induction system accepts three inputs, a set of positive examples $P$, a set of negative examples $N$, and a knowledge base (or ontology) $K$, all expressed as description logic theories, and where we have $x$ occurring as instances (constants) in $K$ for all $x\in P\cup N$. It then returns description logic class expressions $E$ such that $K\models E(p)$ for all $p\in P$ and $K\not\models E(q)$ for all $q\in N$. If no such class expressions exist, then it returns approximations for $E$ together with a number of accuracy measures. In this paper, for scalability reasons, we use the heuristic concept induction system ECII \cite{DBLP:conf/aaai/SarkerH19} together with a background knowledge base that consists only of a class hierarchy, however with approximately 2 million classes, as presented in \cite{DBLP:conf/kgswc/SarkerSHZNMJRA20}. Given a hidden neuron, $P$ is a set of  inputs to the deep learning system that activate the neuron, and $N$ is a set of inputs that do not activate the neuron. Inputs are annotated with classes from the background knowledge for concept induction, however these annotations and the background knowledge are not part of the input to the deep learning system.

As we will see below, this approach is able to provide meaningful explanations for hidden neuron activation.

The rest of this paper is organized as follows. Section \ref{relatedWork} discusses relevant work in the filed of generating explanations using knowledge graph. Sections \ref{researchMethod} present our study design and Section \ref{resultsDiscussion} discusses the results of our study along with the findings and their implications. Finally, Section \ref{conclusion} sums up the paper and proposes some possibilities for future research.

\section{Related Work}
\label{relatedWork}
Explainable AI has been intensively studied since the 1970s \cite{mueller2019explanation}; and the model's explainability can be translated in many ways - interpretable, understandable, justified, and evaluable.

The segment of explainable AI methods focuses on interpreting the inner workings of black box models, such as identifying input features by training explanation networks that generate human-readable explanations \cite{hendricks2016generating} or create models alternatives to summarize the behavior of a complex network \cite{ribeiro2016should}. Other approaches include such as the use of salience maps where the explanations summarize the contribution of each pixel to predictions \cite{bach2015pixel} or visual cues \cite{xu2015show,selvaraju2017grad} or counterfactuals \cite{byrne2019counterfactuals}.


The literature demonstrates that combinations of neurons can encode meaningful and insightful information \cite{kim2018interpretability,bau2017network}.
Justifying the result of a neural network requires a defined language that incorporate elements of reasoning that use knowledge bases to create human-understandable, yet unbiased explanations\cite{doran2017does}.


Knowledge graphs and the structured web represent a valuable form of machine -- readable, domain -- specific knowledge; available connected datasets can serve as a knowledge base for an AI system to explain its decisions to its users in a better way. The Web Ontology Language (OWL) provides a basis for verbose descriptions of
entities and their relationships through description logics \cite{allemang2011semantic}. Deep deductive reasoning can be described as one of generating complex description logic class expressions over the knowledge graph and is based on rich concept hierarchies that play an
important role in generating human -- readable satisfactory explanations. We briefly discussed some recent works  doing logical reasoning using deep networks. 

\cite{zhou2018interpretable,kim2018interpretability,zhou2018interpreting}, methods have been proposed and demonstrated that adding semantic annotations to label objects that activate neurons in the hidden layers of common CNN architectures provides human-readable explanations. Nonetheless, these approaches need to improve in terms of producing deeper explanations generated over more expressive background knowledge.
\cite{procko2022exploration} follows the effort of \cite{sarker2017explaining}, by semi-automating the DL Learner tool, which provides explanations to ML algorithms using semantic background knowledge. However, while DL-Learner is a very useful system in producing theoretically correct results has significant performance issues in some scenarios, such as a single run of DL-Learner can easily take over two hours; in contrast the scenario easily necessitates thousands of such runs.

The main motivation of the proposed work is to automate the assignment of human-interpretable explanations for the activations of neurons in the hidden -- dense layer of CNNs; Using Wikipedia's rich class hierarchy of around 2 million classes with an improved concept induction approach (in terms of running time by 1-2 orders of magnitude while maintaining accuracy of results products) known as ECII.

\section{Research Method}
\label{researchMethod}
This work includes the implementation of explaining the activation pattern of neurons in hidden layers of CNN i.e. dense layer in this case, using Resnet50V2 architecture and ECII -- concept induction explanation generation algorithm. We also tested other architectures to achieve better accuracy and found that Resnet50V2 gives the highest accuracy. The subsections discuss the steps followed for implementing the system in a more detailed manner.

\begin{figure*}[t]
\begin{multicols*}{2}
\begin{subfigure}{0.5\textwidth}
\centering
\includegraphics[width=.2\textwidth]{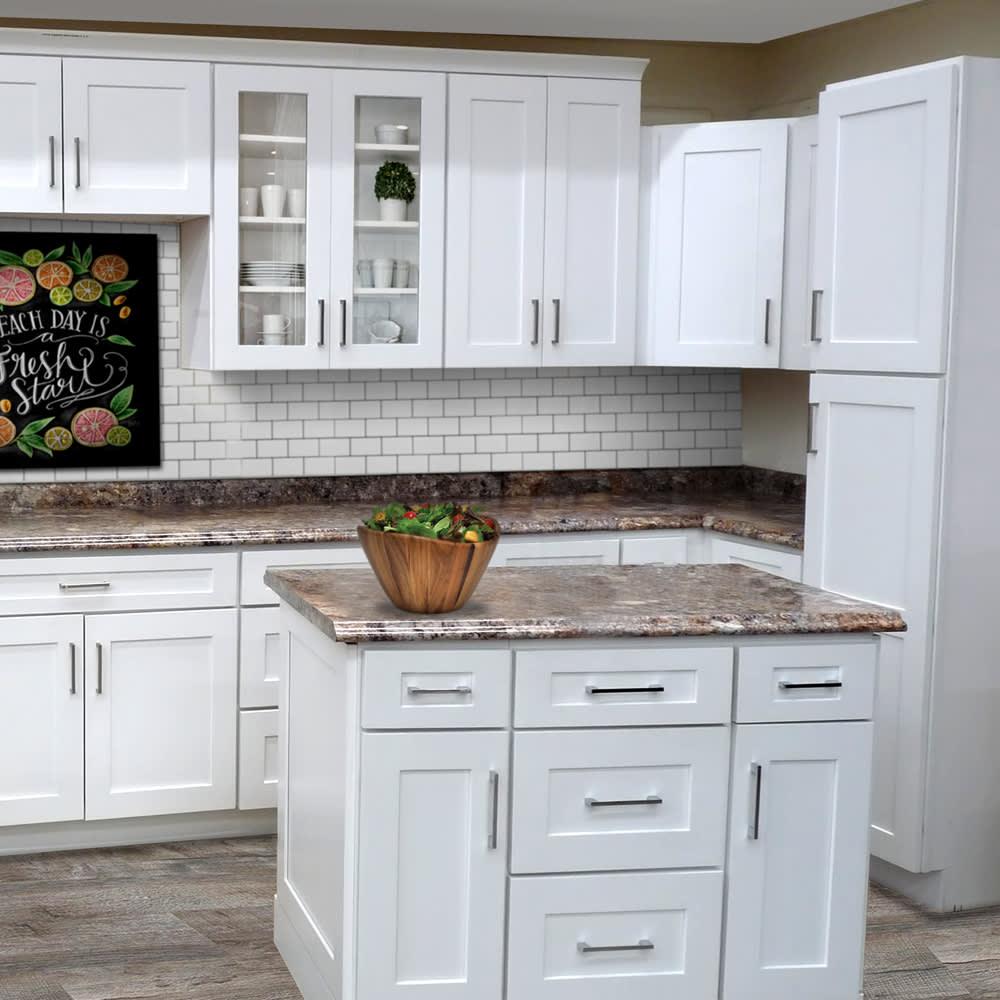}\quad
\includegraphics[width=.3\textwidth]{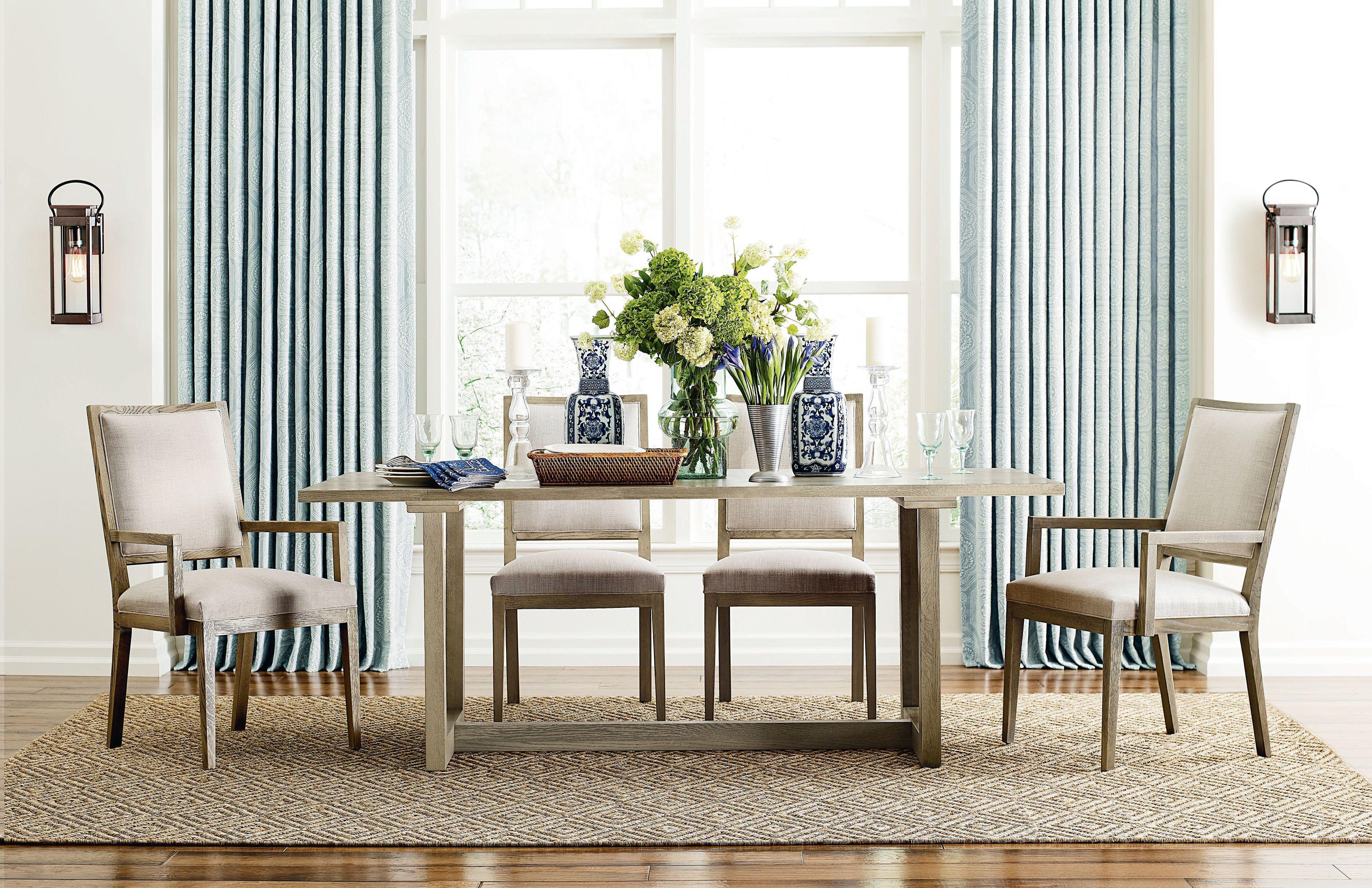}\quad
\includegraphics[width=.2\textwidth]{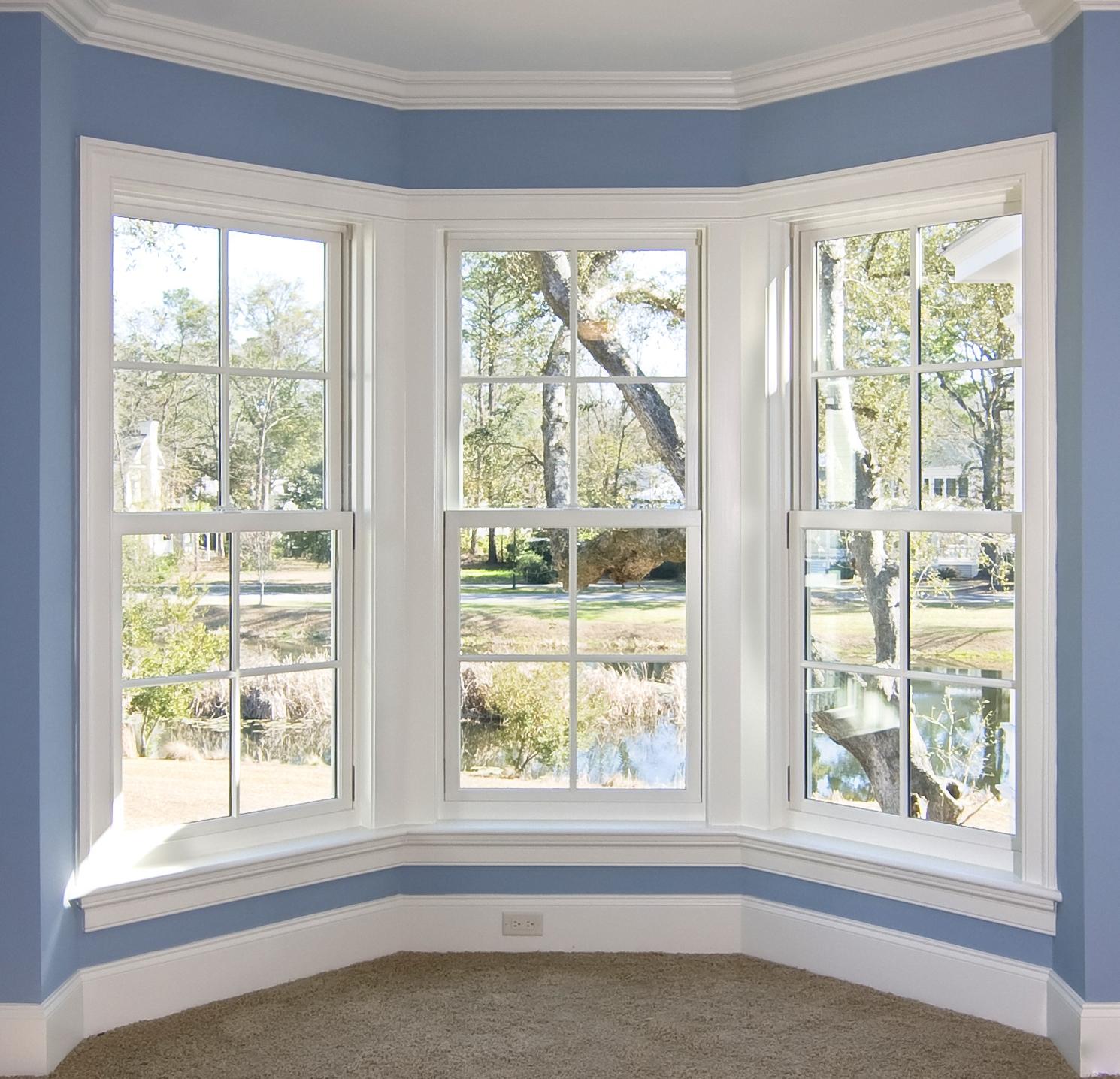}\quad

\medskip

\includegraphics[width=.2\textwidth]{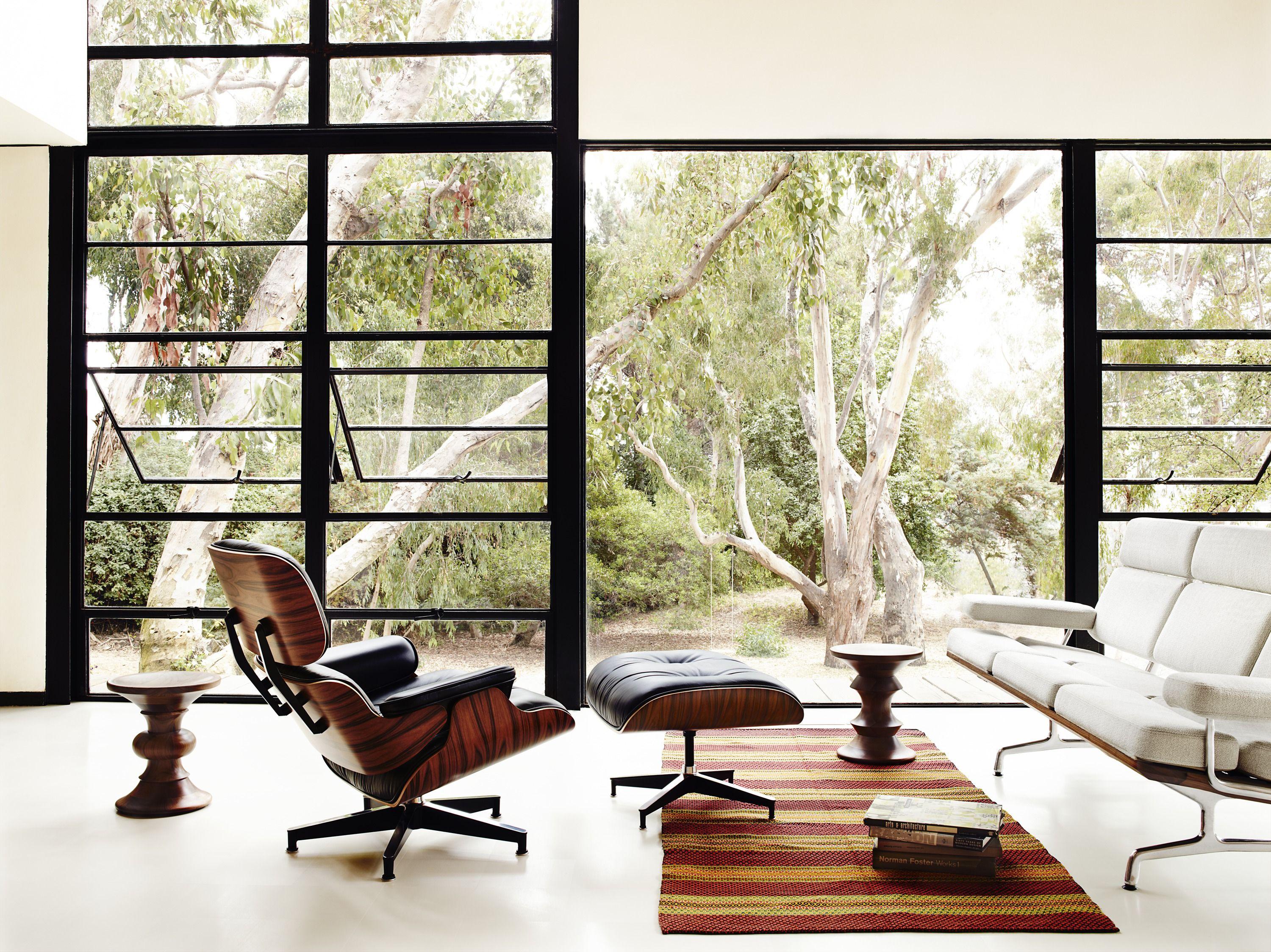}\quad
\includegraphics[width=.2\textwidth]{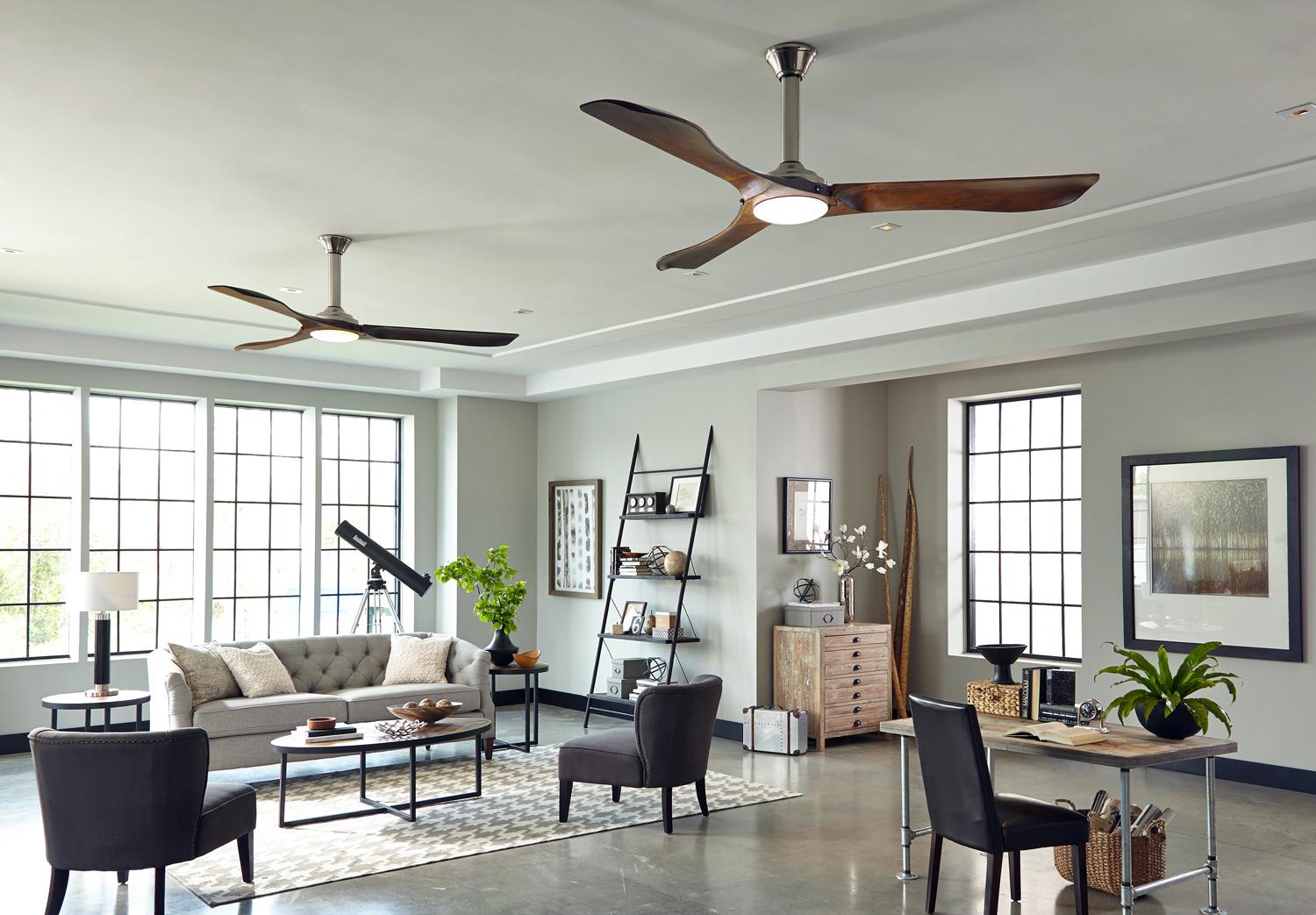}\quad
\includegraphics[width=.2\textwidth]{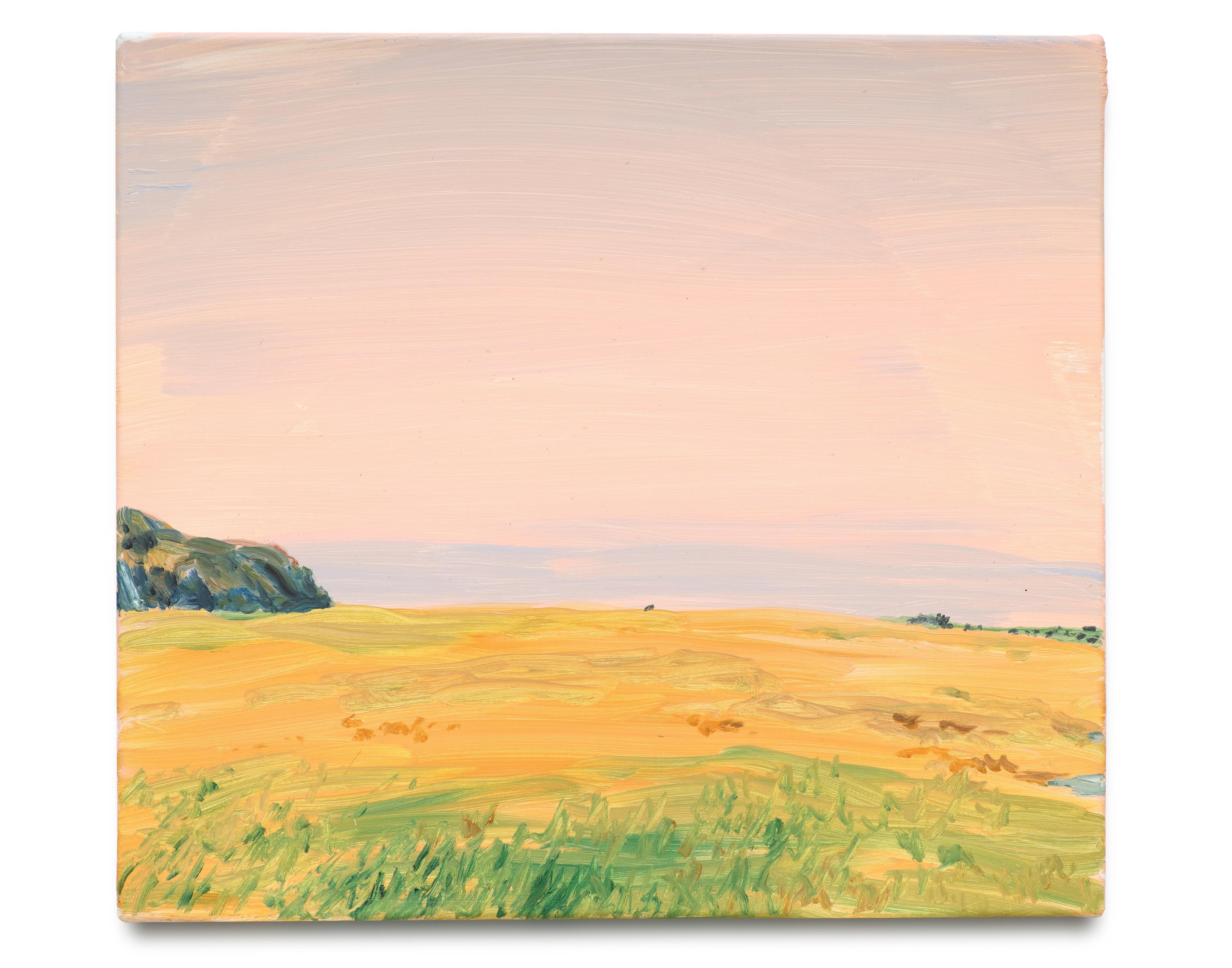}\quad
\includegraphics[width=.2\textwidth]{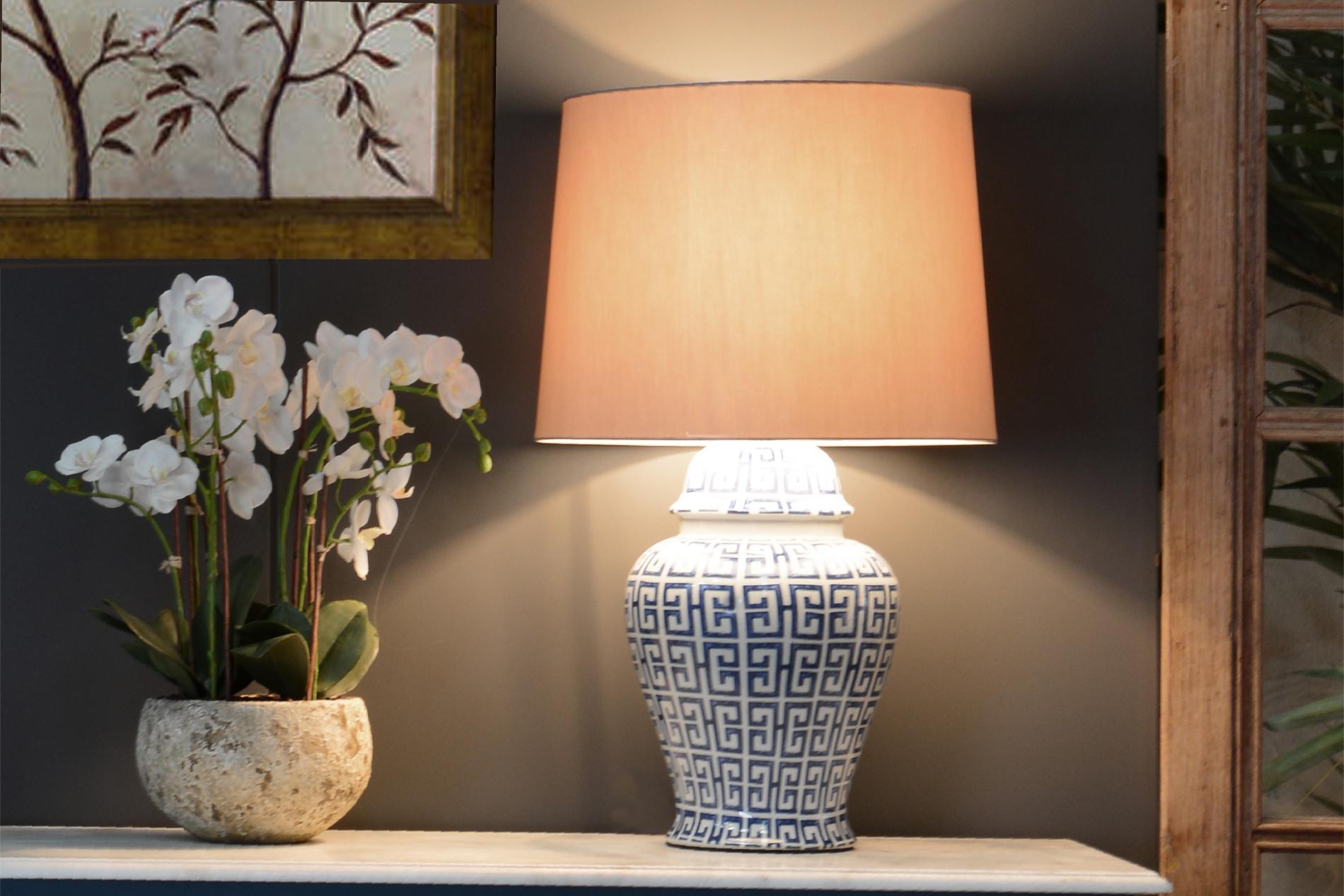}

\caption{Examples of images collected for neuron-5 from google using the lists of concepts for Case-1.}
\label{pics:neuron5dataset_1}
\end{subfigure}

\begin{subfigure}{0.45\textwidth}
\centering
\includegraphics[width=.9\textwidth]{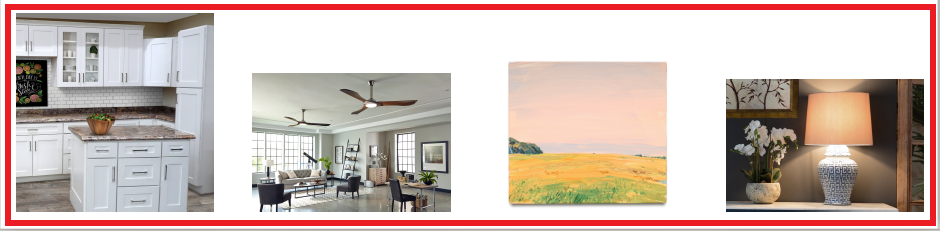}
\caption{Images that didn't activate the neuron-5 for Case-1.}
\label{pics:neuron5_1_ntActivated}
\end{subfigure}

\begin{subfigure}{0.45\textwidth}
\centering
\includegraphics[width=.9\textwidth]{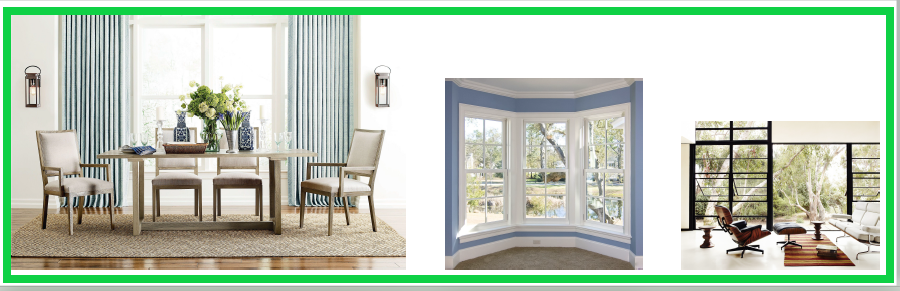}\quad
\caption{Images that activate the neuron-5 for Case-1.}
\label{pics:neuron5_1_Activated}
\end{subfigure}
\end{multicols*}
\caption{Case - I}
\end{figure*}

\subsection{Training Convolutional Neural Network}
\label{TrainingCNN}
\subsubsection{Dataset}
\label{dataset}
1) The ADE20K \cite{zhou2019semantic} semantic segmentation dataset from the Massachusetts Institute of Technology contains more than 27K scene-based images from the SUN and Places databases, extensively annotated with pixel-level objects and object part labels. There are 150 semantic categories including sky, road, grass, and discrete objects like person, car, and bed. The current version of the dataset contains the following:
\begin{itemize}
\item 25,574 for training and 2,000 for testing from 365 scenes.
\item 707,868 unique object along with their WordNet definition and hierarchy.
\item 193,238 parts of annotated objects and parts of parts.
\item Polygon annotations with attributes, annotation time, and depth order.
\end{itemize}
We only considered the subset of scenes in the ADE20k Dataset; the classes that were considered for this work are ten classes with the highest number of images -- bathroom, bedroom, building facade, conference room, dining room, highway, kitchen, living room, skyscraper, and street.

2) For verification purposes of the activation pattern of each neuron in corresponding to identified concepts for that respective neuron, we used Google images -- simply because the system should be easy to use for any user. It should be able to detect concepts and give us the reasoning for its classification category of any random image from the largest crawling search engine. 

\begin{figure*}[t]
\begin{multicols*}{2}
\begin{subfigure}{0.5\textwidth}
\centering
\includegraphics[width=.3\textwidth]{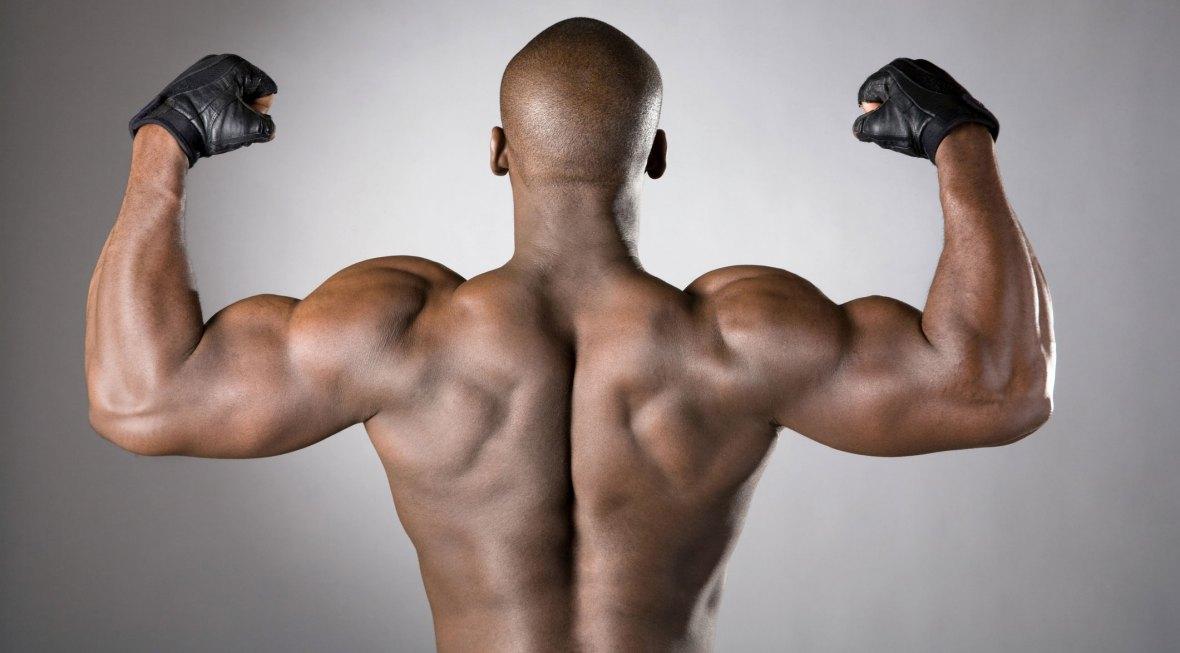}\quad
\includegraphics[width=.3\textwidth]{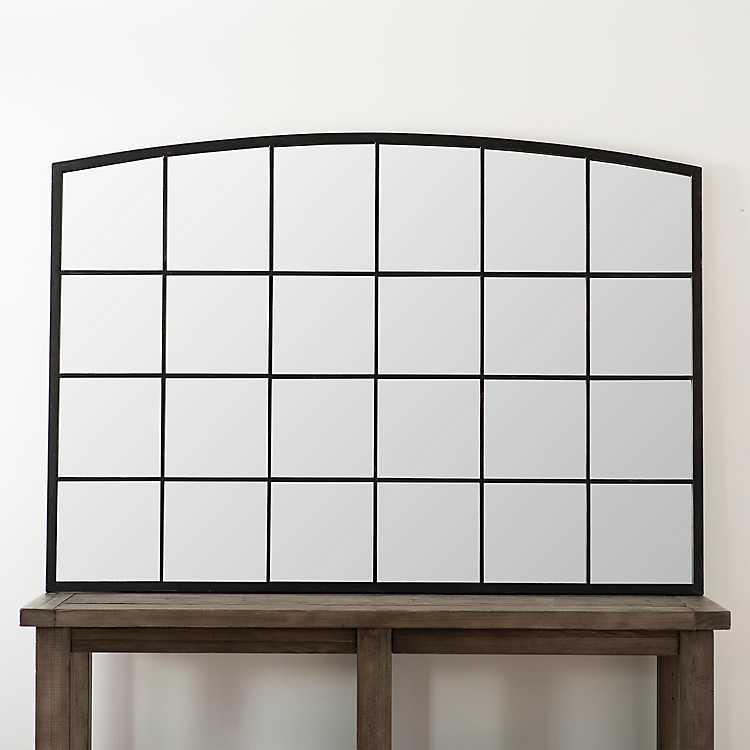}\quad
\includegraphics[width=.3\textwidth]{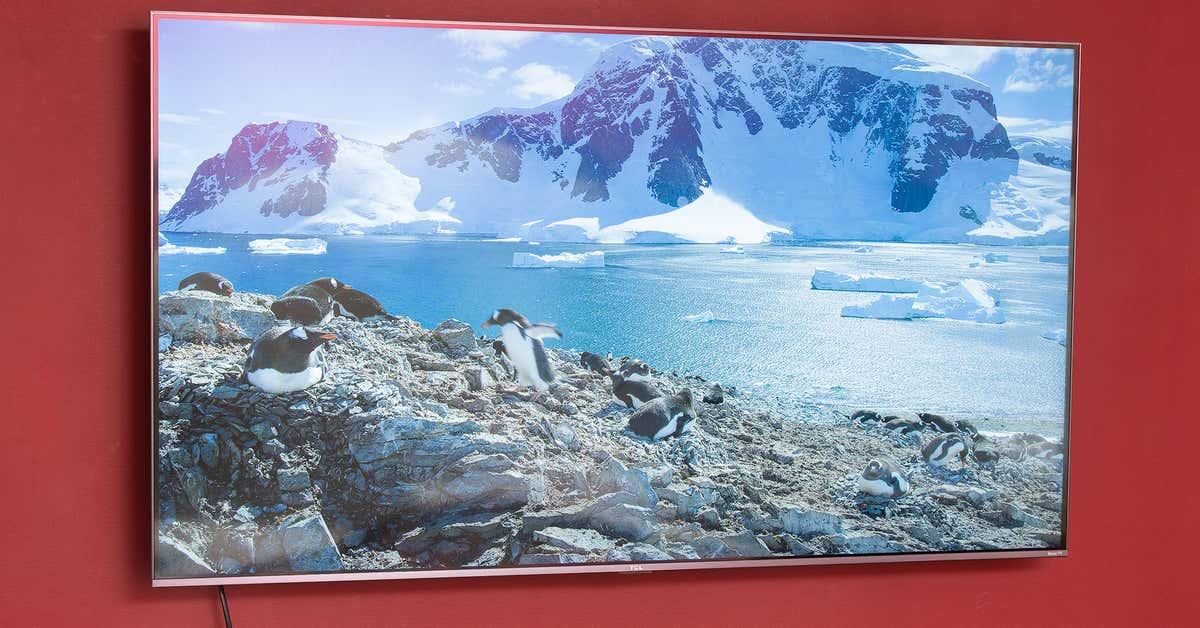}
\medskip
\includegraphics[width=.3\textwidth]{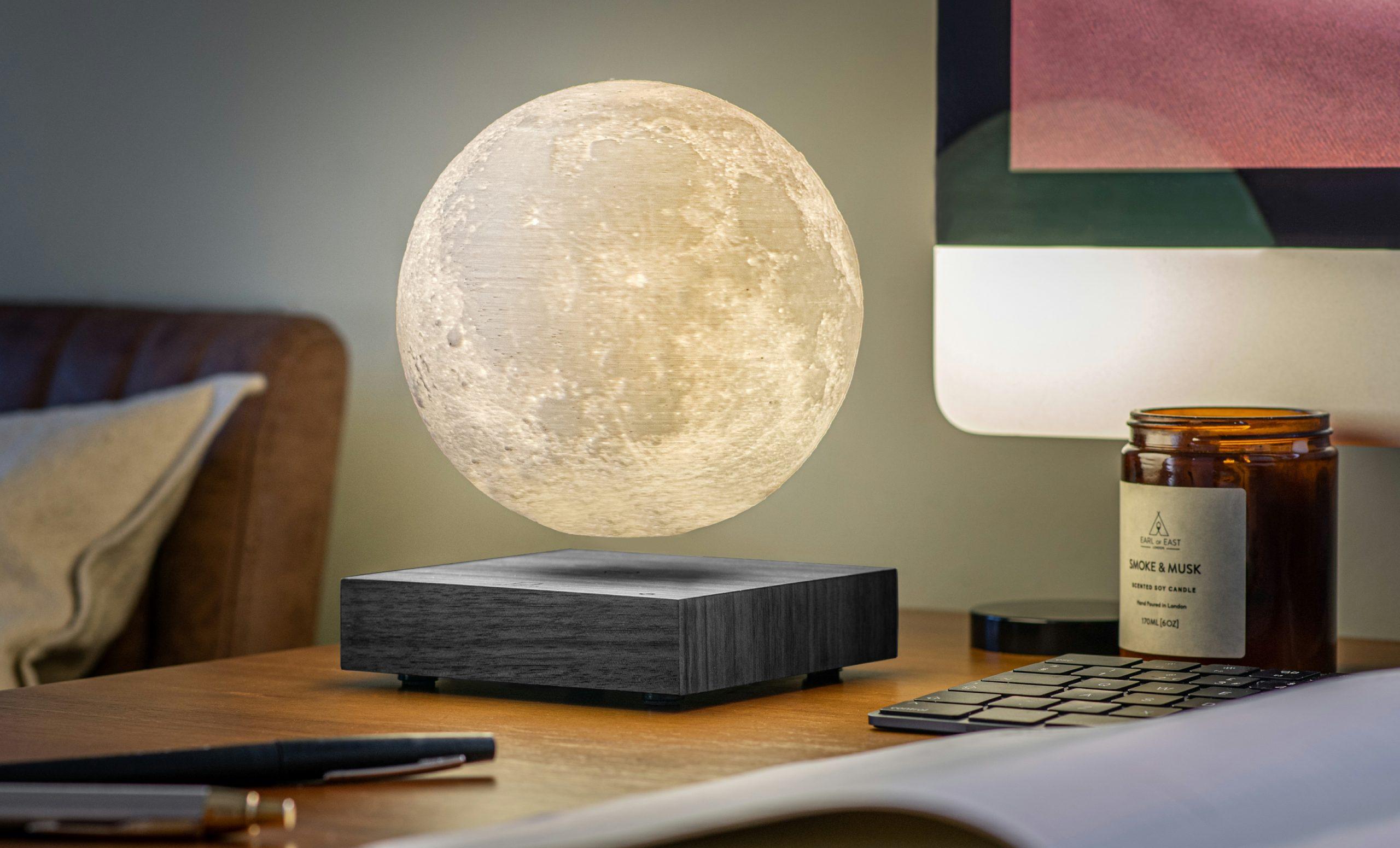}\quad
\includegraphics[width=.3\textwidth]{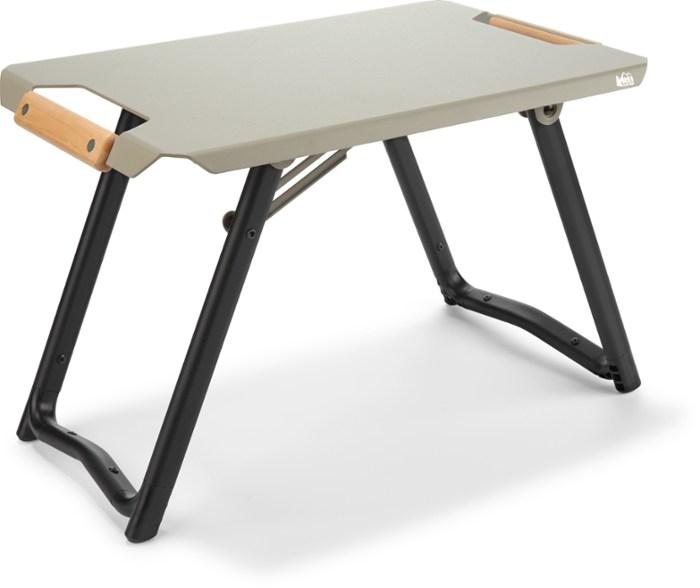}\quad
\includegraphics[width=.3\textwidth]{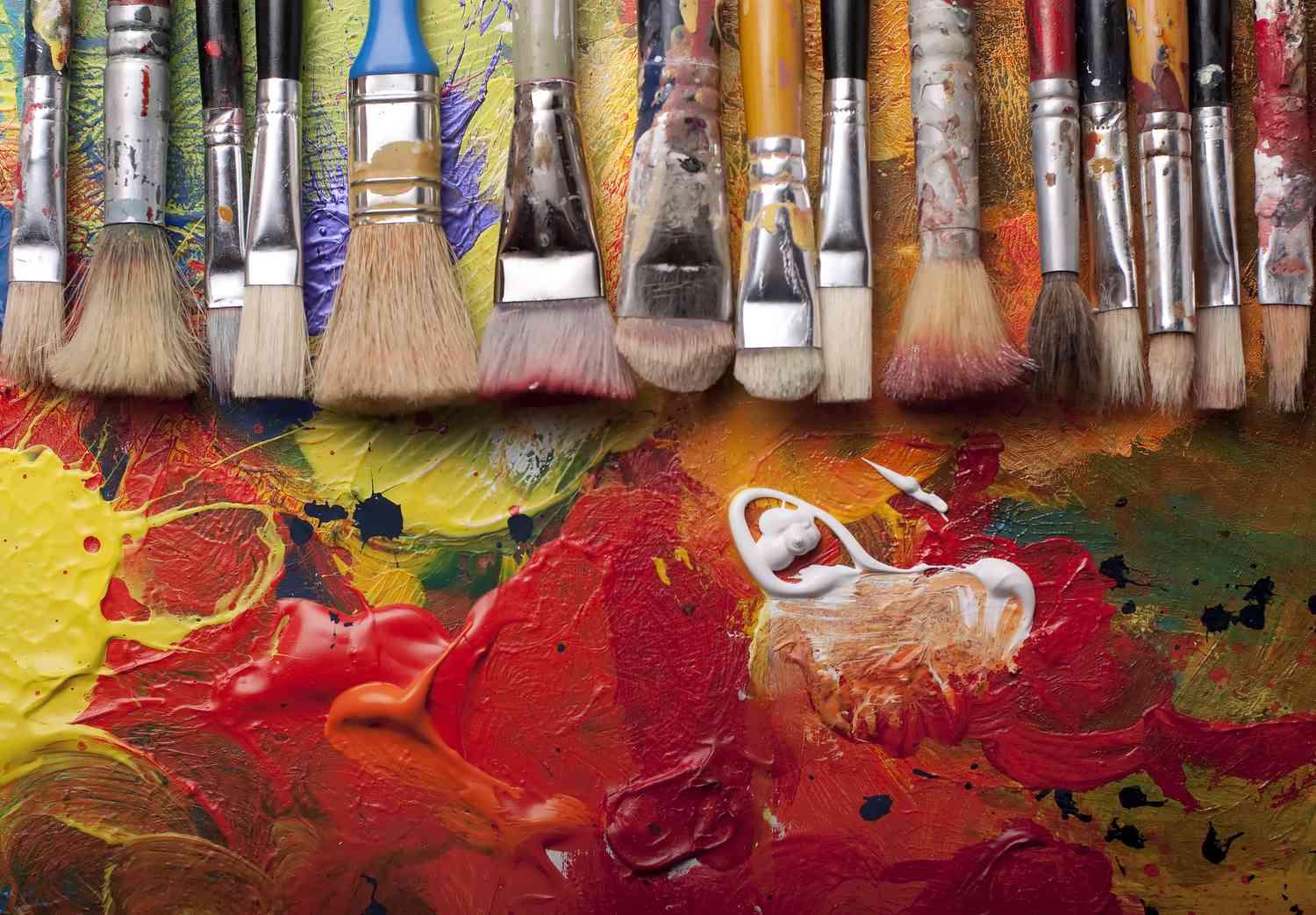}
\caption{Examples of images collected for neuron-5 from google using the lists of concepts for Case-2.}
\label{pics:neuron5dataset_2}
\end{subfigure}

\begin{subfigure}{0.45\textwidth}
\centering
\includegraphics[width=.9\textwidth]{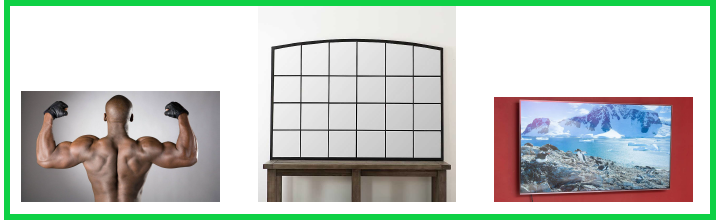}\quad
\caption{Images that activate the neuron-5 for Case-2.}
\label{pics:neuron5_2_activations}
\end{subfigure}

\begin{subfigure}{0.45\textwidth}
\centering
\includegraphics[width=.9\textwidth]{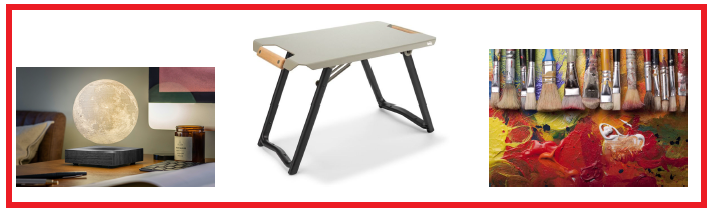}\quad
\caption{Images that didn't activate the neuron-5 for Case-2.}
\label{pics:neuron5_2_ntactivations}
\end{subfigure}
\end{multicols*}
\caption{Case - II}
\end{figure*}

\subsubsection {Tested Networks}
\label{testednetworks}

We analyzed many Convolutional neural network (CNN) architectures to achieve better and higher accuracy such as Vgg16 \cite{simonyan2014very}, InceptionV3 \cite{szegedy2016rethinking}; in Resnet we tried different versions like -- Resnet50, and Resnet50V2, Resnet101, Resnet152V2 architecture \cite{He2015,he2016identity}.




Each neural network was fine-tuned with a dataset of 6187 images (training and validation set) of size 224*224 for 20 epochs to classify images into 10 scene categories using the ADE20K dataset. The optimization algorithm used was Adam, with a categorical cross-entropy loss function and a learning rate of 0.001. The accuracy achieved by each architecture along with validation accuracy is summarized in table \ref{accuracy_of_networks}.
\begin{table}
    \centering
    \begin{tabular}{lrr}
        \toprule
        Architectures  & Training acc & Validation acc \\
        \midrule
        Vgg16          & 80.05\%      & 46.22\%        \\
        InceptionV3    & 89.02\%      & 51.43\%         \\
        Resnet50       & 35.01\%      & 26.56\%        \\
\textbf{Resnet50V2}    &\textbf{92.47\%}      &\textbf{87.50\%}  \\
        Resnet101      & 53.97\%      & 53.57\%        \\
        Resnet152V2    & 94.53\%      & 51.04\%        \\
        \bottomrule
    \end{tabular}
    \caption{Performance of different architectures on ADE20K dataset}
    \label{accuracy_of_networks}
\end{table}

Clearly, ResNet50V2 achieved the highest accuracy -- 92.47\% on the training dataset and 87.50\% on the validation dataset, proving to be the best network out of all.

\subsubsection {Activations of Trained Network}
\label{activationsTrainedNw}
We tested the Resnet50V2 with 1370 images and retrieved the activations of the dense layer, i.e., the layer before the output layer. Though technically, the layer before the output layer is the dropout layer, we chose not to analyze the activations of the dropout layer since the dropout layer is a mask that negates the contribution of some neurons towards the next layer and leaves all others unmodified.

The activations of 1370 images for the dense layer comprise 64 neurons contributing to the final decision of classifying each image as one of 10 classes. 

\subsubsection {Candidate Set of Neurons}
\label{candidadateSet}
Next, out of 64 neurons, we chose some candidate sets of neurons based on the following criteria -- only the neuron having more than 50\% of activation values $>$ 0 i.e., the neuron should have at least 680 values (= 1370/2, 1370 being total images) that are greater than 0. Choosing such neurons would simply mean that these are frequently activated nodes, which would be a good choice to analyze before exploring any other possibilities.
Following the idea, the neurons selected for analysis were neuron numbers -- 4, 5, 6, 7, 9, 11, 12, 13, 15, 16, 22, 23, 27, 29, 34, 35, 36, 37, 39, 45, 52, 54, 55, 56, 58, 59, 60, 62, 63.

\subsubsection {ECII - Preliminaries}
\label{eciiprelim}
As mentioned concept induction is an explanation generation algorithm over description logic which takes in three inputs -- a positive set of images, a negative set of images and a knowledge base. For our approach, we use ECII --improved on DL Learner by the magnitude of order 2. 

For a given neuron, a positive set of images that activates the said neuron, and a negative set of images that do not activate the given neuron. How do we decide that an image activates a neuron and therefore that image is positive, and in the same way for negative set of images.
To decide on activation, we considered and analyzed -- a threshold value around the activation values with the following three different criteria:-
\begin{itemize}
\item CASE-I -- positive set will have images with $>=$ 50\% activation of the highest value, lets say if the highest activation value is 12 for neuron\_x then all images (1370, is the total number) having an activation value of 6 or more than 6 will be positive set and so negative set will have images that were $<$ 50\% i.e all the images with less than 6 including 0.
\item CASE-II -- positive set will have images with $>=$ 50\% activation of the highest activation value and negative set as the images that were just zero, i.e excluding images that are 0 $<$ images $<$ 50\%.
\item CASE-III -- positive set will have images with anything $>$ zero i.e this will include all images that are 0 $<$ images $<=$ highest value and negative set as the images that were just zero, i.e excluding images that are 0 $<$ images.
\end{itemize}

For the knowledge base, we mapped all the 1370 images with Wikipedia's rich class hierarchy of 2 million classes.

\subsubsection {ECII - Analysis}
Now that we have a knowledge base and a set of positive and negative images based on three cases, we run ECII with all three inputs from each case defined above for all chosen candidate sets of the neurons.
ECII returns a list of class expressions such that it best describes the positive set of images while excluding all negative images, sorted by coverage score.

Coverage score can be formulated using the following formula:
$$coverage(E) = \frac{|P\cap Z1 | + |N\cap Z2|}{|P\cup N|}$$
Where,\\
\begin{center}{
$Z1 = K\models E(p)$ for all $p\in P$, \\
$Z2 = K\not\models E(n)$ for all $n\in N$,\\
$P$ is the set of all positive instances,\\ $N$ is the set of all negative instances, and \\$K$ is the knowledge base provided to ECII as input.} 
\end{center}

We chose to look at the first 50 expressions out of all returned by ECII in text format, simply because the list of expressions could have many duplicate concepts.
\begin{example}
An example of the -- explanation ECII came up with looks like 
    
    solution 1 $\exists$ imageContains.((WN\_Table) $\sqcap$ (Bed))
    
    solution 3 $\exists$ imageContains.(:WN\_Table)
\end{example}

indicating the presence of a table and bed in one of the images from the positive set. We collected all distinctive keywords as concepts(in this case -- Table and Bed) from the list since solutions could have overlapping concepts, resulting in a reduced list of concepts.

This returned list of concepts for each neuron would give us the intuition of what contributes towards the activation of the respective neuron.
\begin{example}
As an example lets see the the list of class expression returned by ECII for neuron 5 and its corresponding reduced list of concepts.

solution 1: $\exists$ :imageContains.(:WN\_Table)\\
solution 2: $\exists$ :imageContains.(:Floor)\\
solution 3: $\exists$ :imageContains.(:WN\_Floor)\\
solution 4: $\exists$ :imageContains.(:WN\_Flooring)\\
solution 5: $\exists$ :imageContains.(:Window)\\
solution 6: $\exists$ :imageContains.(:WN\_Window)\\
solution 7: $\exists$ :imageContains.((:WN\_Flooring) $\sqcap$ (:Window))\\
solution 8: $\exists$ :imageContains.((:Window) $\sqcap$ (:Floor))\\
solution 9: $\exists$ :imageContains.((:WN\_Flooring) $\sqcap$ (:Floor))\\
solution 10: $\exists$ :imageContains.((:Ceiling) $\sqcap$ (:WN\_Table))\\
solution 11: $\exists$ :imageContains.(:Ceiling)\\
solution 12: $\exists$ :imageContains.(:WN\_Ceiling)\\
solution 13: $\exists$ :imageContains.(:WN\_Windowpane)\\
solution 14: $\exists$ :imageContains.(:WN\_Leg)\\
solution 15: $\exists$ :imageContains.(:Picture)\\
solution 16: $\exists$ :imageContains.(:WN\_Painting)\\
solution 17: $\exists$ :imageContains.(:WN\_Picture)\\
solution 18: $\exists$ :imageContains.(:Leg)\\
solution 19: $\exists$ :imageContains.((:WN\_Table) $\sqcap$ (:Leg))\\
solution 20: $\exists$ :imageContains.((:WN\_Painting) $\sqcap$ (:WN\_Ceiling))\\
solution 21: $\exists$ :imageContains.((:WN\_Leg) $\sqcap$ (:WN\_Window))\\
solution 22: $\exists$ :imageContains.(:Chair)\\
solution 23: $\exists$ :imageContains.(:WN\_Chair)\\
solution 24: $\exists$ :imageContains.(:WN\_Lamp)\\
solution 25: $\exists$ :imageContains.((:WN\_Lamp) $\sqcap$ (:WN\_Floor))\\
solution 26: $\exists$ :imageContains.((:WN\_Windowpane) $\sqcap$ (:WN\_Painting))\\
solution 27: $\exists$ :imageContains.(:Back)\\
solution 28: $\exists$ :imageContains.(:WN\_Back)\\
solution 29: $\exists$ :imageContains.((:Back) $\sqcap$ (:WN\_Flooring))\\
solution 30: $\exists$ :imageContains.((:WN\_Floor) $\sqcap$ (:WN\_Back))\\
solution 31: $\exists$ :imageContains.((:WN\_Windowpane) $\sqcap$ (:WN\_Ceiling))\\
solution 32: $\exists$ :imageContains.((:Ceiling) $\sqcap$ (:Leg))\\
solution 33: $\exists$ :imageContains.((:Floor) $\sqcap$ (:Table))\\
solution 34: $\exists$ :imageContains.(:Table)\\
solution 35: $\exists$ :imageContains.((:WN\_Back) $\sqcap$ (:WN\_Windowpane))\\
solution 36: $\exists$ :imageContains.((:Chair) $\sqcap$ (:Ceiling))\\
solution 37: $\exists$ :imageContains.(:Arm)\\
solution 38: $\exists$ :imageContains.(:WN\_Arm)\\
solution 39: $\exists$ :imageContains.((:WN\_Window) $\sqcap$ (:WN\_Lamp))\\
solution 40: $\exists$ :imageContains.((:Back) $\sqcap$ (:Window))\\
solution 41: $\exists$ :imageContains.((:WN\_Floor) $\sqcap$ (:WN\_Windowpane))\\
solution 42: $\exists$ :imageContains.((:Back) $\sqcap$ (:Floor))\\
solution 43: $\exists$ :imageContains.((:WN\_Window) $\sqcap$ (:WN\_Floor))\\
solution 44: $\exists$ :imageContains.((:Chair) $\sqcap$ (:WN\_Table))\\
solution 45: $\exists$ :imageContains.(:Top)\\
solution 46: $\exists$ :imageContains.(:WN\_Top)\\
solution 47: $\exists$ :imageContains.((:Table) $\sqcap$ (:WN\_Chair))\\
solution 48: $\exists$ :imageContains.((:Floor) $\sqcap$ (:WN\_Chair))\\
solution 49: $\exists$ :imageContains.((:Leg) $\sqcap$ (:Picture))\\
solution 50: $\exists$ :imageContains.(:WN\_Cabinet)\\

And after eliminating duplicate concepts from the above class expressions, we get a reduced list of concepts as -- 

\textbf{arm, back, cabinet, ceiling, chair, floor, flooring, lamp, leg, painting, picture, table, top, window, windowpane}
\end{example}

The next step would be to verify the activation of neurons by collecting some more data that is more generic and could serve as solid verification and test it through the model and see if we get the same activations for the neurons; we collected google images corresponding to the resultant keywords of the list for each neuron. In this case -- collect images of arm, back, cabinet, ceiling, chair, floor, flooring, lamp, leg, painting, picture, table, top, window, and windowpane from the google search engine. 

\subsubsection {Collection of Google Images}
 We used a python script to download Google images for each keyword in the list. For each keyword, it collects the first 200 images that appear in the google search. After that, we manually checked for duplicates and removed them. After cleaning the duplicates we have at least 140 images for each keyword. For example, for the keyword, 'base' google search comes up with all kinds of images including bed frames to military bases. However, some search for keywords like 'edifice' collects images of a particular model of watch named edifice, which in this case is not what we wanted. But we take the google results as it appears and evaluate our model on them.     
\subsubsection {Activations on Google Images}
Once we have the dataset from google ready for all neurons (29 being total as chosen candidate set), we test this new dataset for each neuron through our trained model -- Resnet50V2 and get the activations of the dense layer.

\begin{figure*}[t]
\begin{multicols*}{2}
\begin{subfigure}{0.5\textwidth}
\centering
\includegraphics[width=.3\textwidth,height = 2cm]
{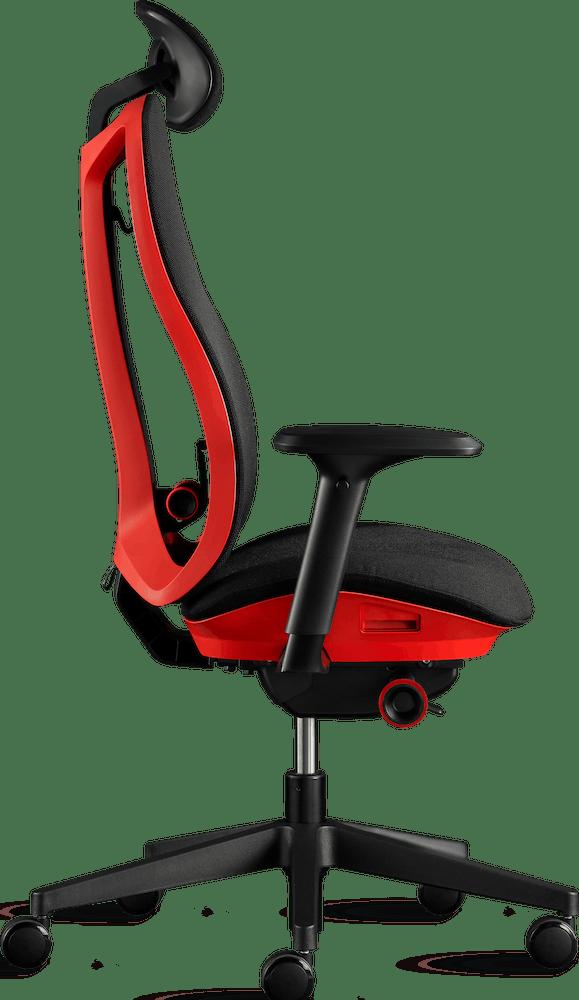}\quad
\includegraphics[width=.3\textwidth]{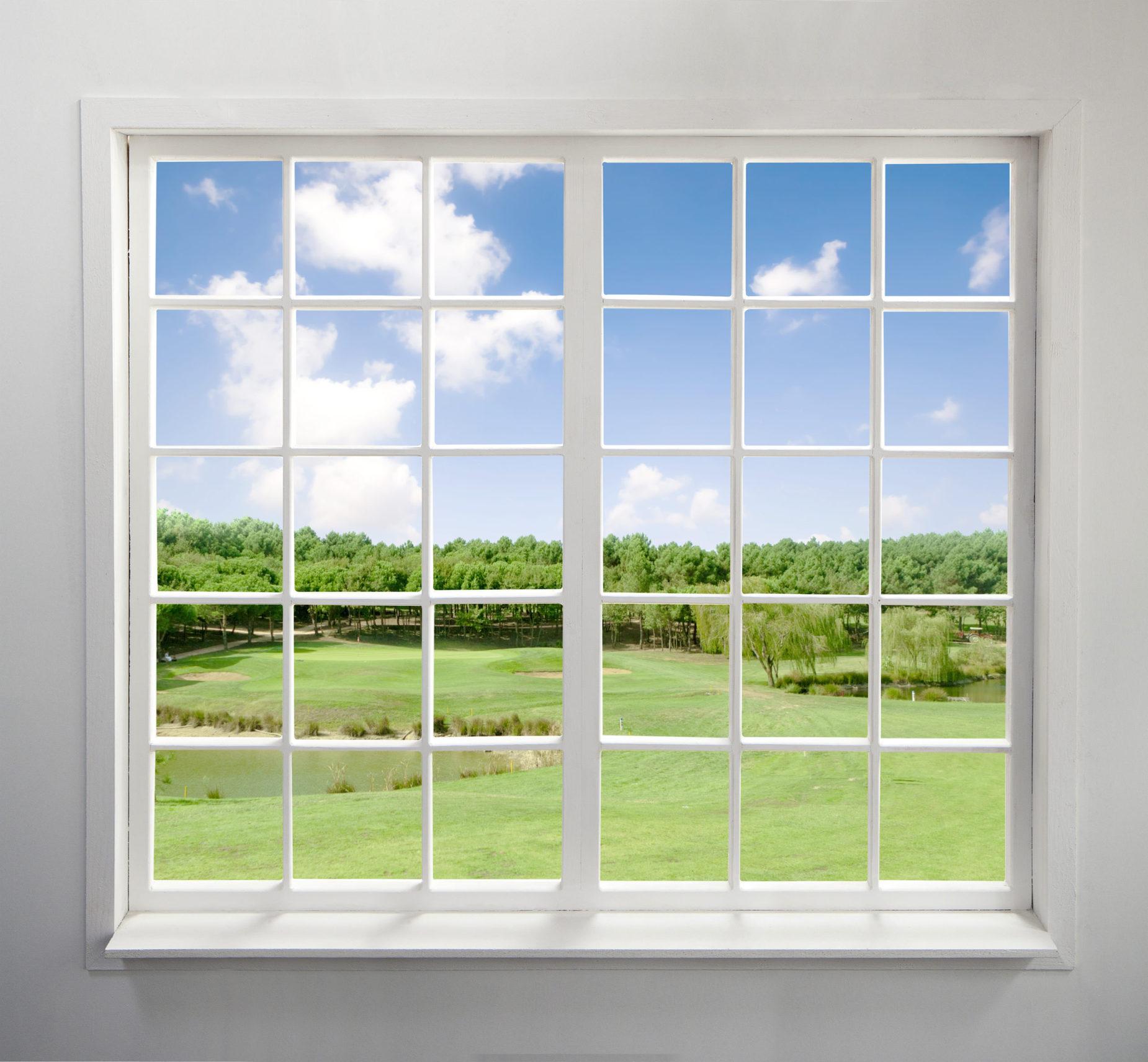}\quad
\includegraphics[width=.3\textwidth]{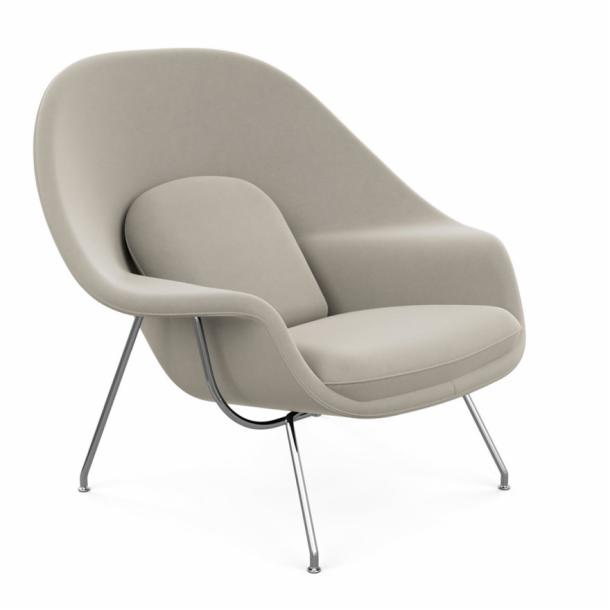}\quad

\medskip

\includegraphics[width=.22\textwidth,height = 2cm]{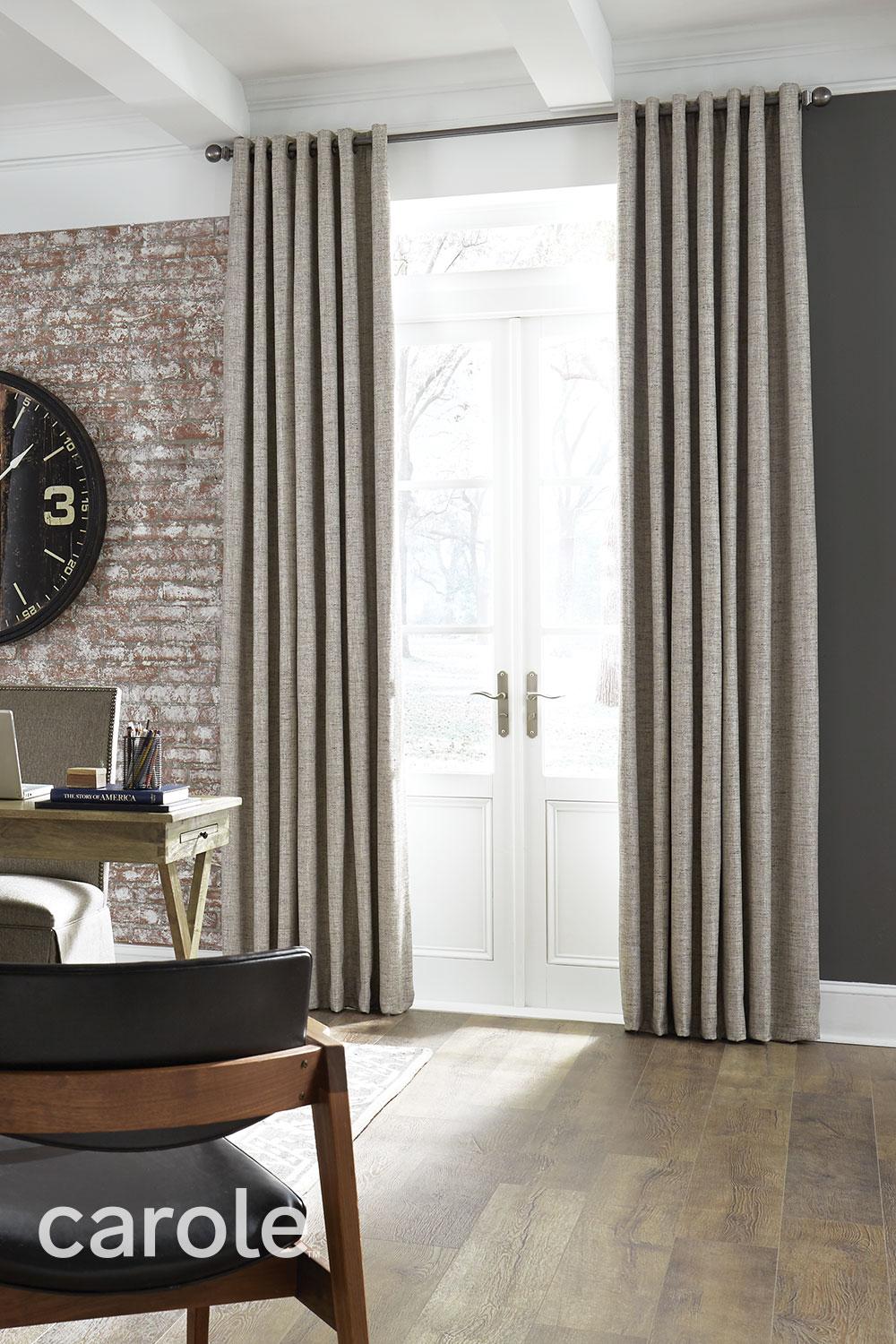}
\includegraphics[width=.22\textwidth]{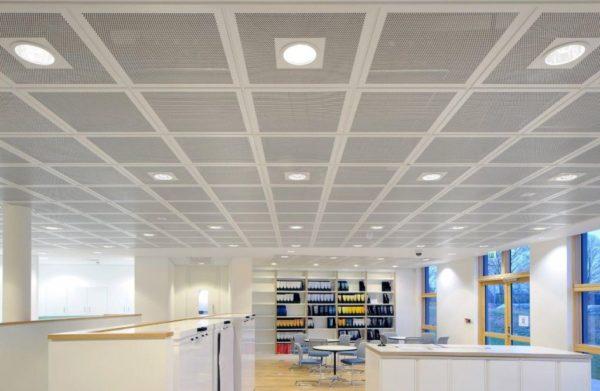}\quad
\includegraphics[width=.22\textwidth]{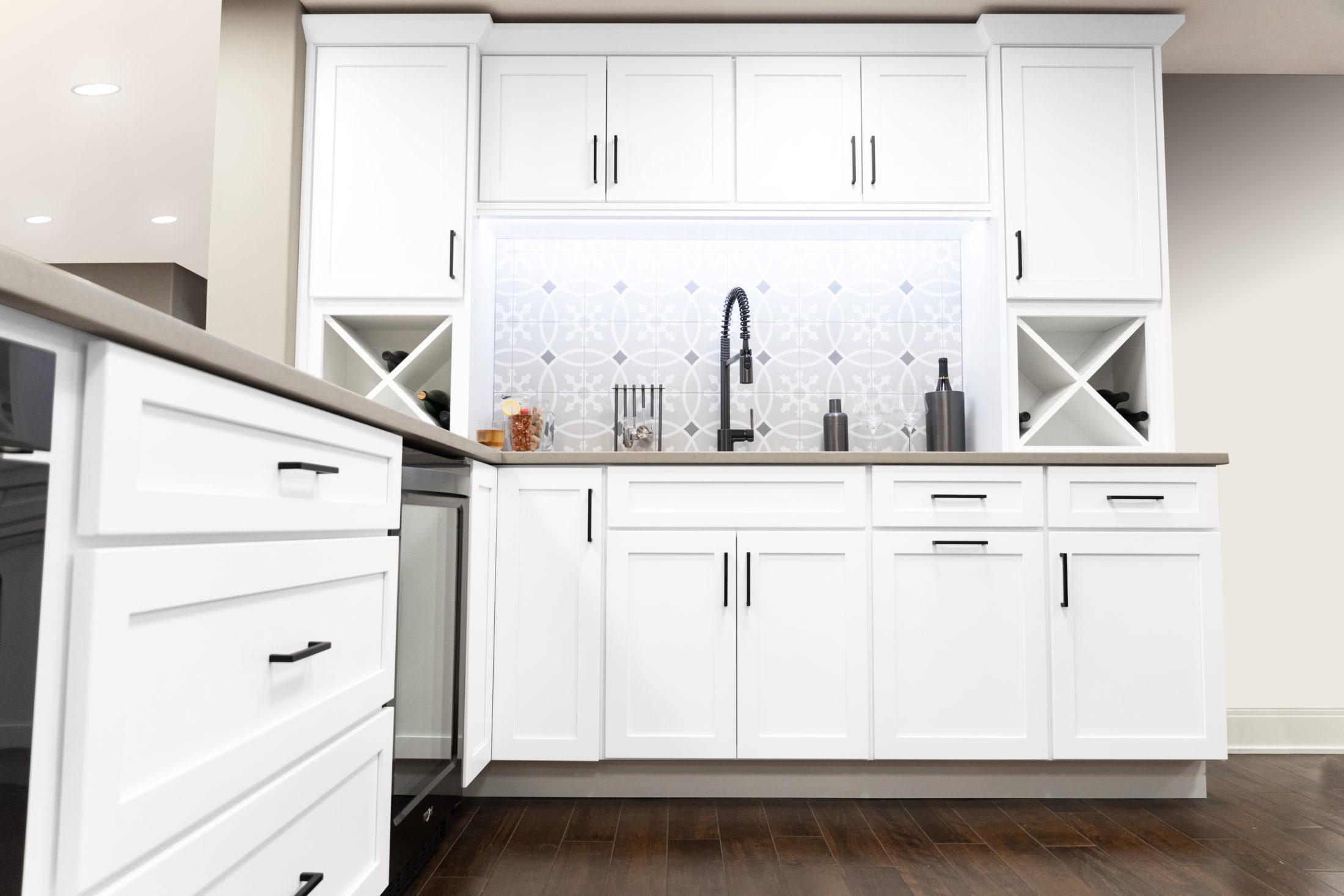}\quad
\includegraphics[width=.22\textwidth]{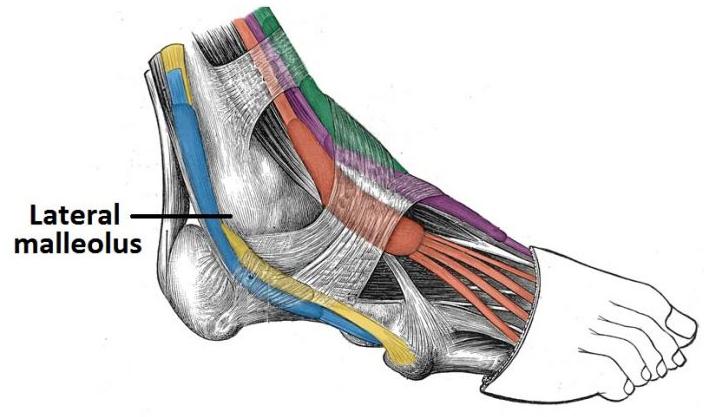}

\caption{Examples of images collected for neuron-5 from google using the lists of concepts for Case-3.}
\label{pics:neuron5dataset_3}
\end{subfigure}

\begin{subfigure}{0.45\textwidth}
\centering

\includegraphics[width=.9\textwidth]{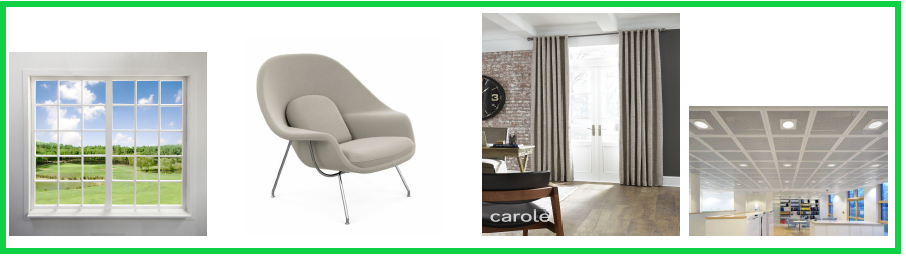}

\caption{Images that activate the neuron-5 for Case-3.}
\label{pics:neuron5_3Activations}
\end{subfigure}

\begin{subfigure}{0.45\textwidth}
\centering

\includegraphics[width=.9\textwidth,height = 2cm]
{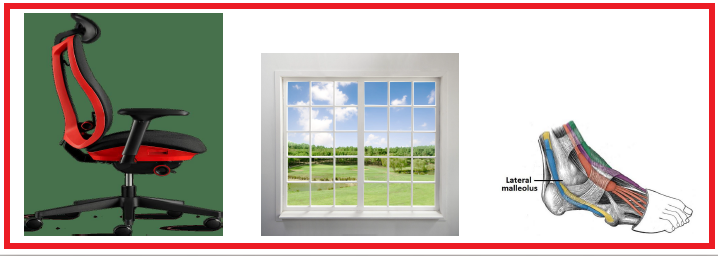}\quad

\caption{Images that didn't activate the neuron-5 for Case-3.}
\label{pics:neuron5_3ntActivations}
\end{subfigure}
\end{multicols*}
\caption{Case - III}
\end{figure*}

\section{Results and Discussion}
\label{resultsDiscussion}
We analyzed three different criteria as mentioned in subsection \ref{eciiprelim} -- ECII-Preliminaries, to see what we could call the best scenario for the activation of the neuron --
\begin{itemize}
\item CASE-I -- positive set will have images with $>=$ 50\% activation of the highest value, and the negative set will have images that were $<$ 50\%.
\item CASE-II -- positive set will have images with $>=$ 50\% activation of the highest activation value and negative set as the images that were just zero.
\item CASE-III -- positive set will have images with anything $>$ zero and negative set as the images that were just zero.
\end{itemize}

ECII was run for all neurons taking each case at a time along with Wikipedia as a Knowledge base; in total we did 29*3 = 87 ECII analysis.

From each ECII analysis, we got a list of class expressions sorted by coverage score for the respective neuron -- looked at the first 50 expressions, and reduced it to a shorter list by eliminating any duplicate keywords. These keywords indicate the activation of neurons by the presence of these concepts. Table \ref{tab:listofConcepts} lists the concepts we got from ECII corresponding to each case, representing the neuron's activation for neuron number 5. 
\begin{table}[t]
    \centering
    \begin{tabular}{lrr}
        \toprule
          Case-I & Case-II & Case-III \\
        \midrule
        arm         & arm       & cabinet\\
        back        & back      & ceiling \\
        cabinet	    & cabinet	& chair  \\
        ceiling	    & ceiling	& curtain\\
        chair       & chair	    & cushion\\
        floor	    & floor	    & drapery\\
        flooring    & flooring	& floor\\
        lamp	    & lamp	    & flooring\\
        leg	        & leg	    & lamp\\
        painting	& painting	& leg\\
        picture	    & picture	& painting\\
        table	    & table	    & picture\\
        top	        & wall	    & shade\\
        window	    & windowpane & table\\
        windowpane	&            &table\_lamp\\
                    &            & wall\\
                    &            & windowpane\\
        \bottomrule
    \end{tabular}
    \caption{List of concepts activating neuron\_5 for case I, II, III}
    \label{tab:listofConcepts}
\end{table}

To verify if these concepts actually play a role for neurons in deciding the output for the  network, collected google images corresponding to the reduced list of concepts for each neuron. 

In case of neuron number 5 CASE--I, google images were collected corresponding to \textbf{arm, back, cabinet, ceiling, chair, floor, flooring, lamp, leg, painting, picture, table, top, window, windowpane.}

For CASE--II, google images were collected corresponding to \textbf{ arm, back, cabinet, ceiling, chair, floor, flooring, lamp, leg, painting, picture, table, wall, windowpane.}

For CASE--III, google images were collected corresponding to  \textbf{cabinet, ceiling, chair, curtain, cushion, drapery, floor, flooring, lamp, leg, painting, picture, shade, table, table\_lamp, wall, windowpane.}

Around 200 images were collected for each concept in the list, making a total of around 4000-5000 images for each neuron. In total 4000*87 = 348000 images were collected. 

The new google image dataset was divided into 80-20 ratio and 80\% of them were tested by the trained model for verification and activations of the dense layer (n-1 layer) were analyzed for each neuron. In total there were 87 dense layer activations; each dense layer activation consists of 64 neurons; we only look for the activation value of the desired neuron number. The activation percentage for each neuron is summarized case--wise in table \ref{activationGoogle}. The figure shows the examples of the google image dataset collected for neuron\_5 in each case, along with images that activated the neuron and those that didn't activate the neuron.

\begin{table}
    \centering
    \begin{tabular}{lrrr}
        \toprule
         & Case-I  & Case-II & Case-III \\
        \midrule
        neuron4	    &100\%		&100\%		&100\%\\
        neuron5     &35.01\%	&36.84\%	&38.38\%\\
        neuron6     &0.44\%		&0.40\%		&0.24\%\\
        neuron7     &6.31\%		&7.48\%		&5.71\%\\
        neuron9     &99.90\%	&100\%		&99.97\%\\
        neuron11    &99.00\%	&99.00\%	&99.00\%\\
        neuron12    &95.20\%	&95.20\%	&95.20\%\\
        neuron13    &0.05\%		&0.06\%		&0.05\%\\
        neuron15    &99.93\%	&99.96\%	&100\%\\
        neuron16    &99.94\%	&99.97\%	&99.97\%\\
        neuron22    &37.32\%	&26.00\%	&26.24\%\\
        neuron23    &100\%		&100\%		&100\%\\
        neuron27    &99.64\%	&99.65\%	&99.60\%\\
        neuron29    &0.67\%		&0.67\%		&0.67\%\\
        neuron34    &56.46\%	&56.46\%	&57.58\%\\
        neuron35    &16.32\%	&16.31\%	&9.25\%\\
        neuron36    &0\%		&0\%		&0\%\\
        neuron37    &0\%		&0\%		&0\%\\
        neuron39    &0.24\%		&0.20\%		&0.63\%\\
        neuron45    &0\%		&0.09\%		&0\%\\
        neuron52    &0.18\%		&0.24\%		&0.17\%\\
        neuron54    &0\%		&0\%		&0\%\\
        neuron55    &53.81\%	&47.88\%	&38.36\%\\
        neuron56    &4.16\%		&4.06\%		&2.22\%\\
        neuron58    &1.80\%		&1.80\%		&1.68\%\\
        neuron59    &0\%		&0\%		&0\%\\
        neuron60    &100\%		&99.96\%	&100\%\\
        neuron62    &100\%		&100\%		&100\%\\
        neuron63    &100\%		&100\%		&100\%\\
        \bottomrule
    \end{tabular}
    \caption{Activation percentage for each neuron with Google Images for all three cases.}
    \label{activationGoogle}
\end{table}

Some observations from the table --
\begin{itemize}
\item 11 neurons -- neuron number 4, 9, 11, 12, 15, 16, 23, 27, 60, 62, and 63 got activated by more than 90\% activations in all the three cases. 
\item 10 neurons -- neuron numbers 6, 13, 29, 36, 37, 39, 45, 52, 54, 59 were below 1\% activations in all three cases.
\item the rest activations are in the range of 1 -- 56.52\%.
\end{itemize}

We can say that the criteria we chose for deciding the activation for the positive and negative set of images -- as two inputs for ECII, doesn't have much impact on the activation percentage of the neurons as there is a slight difference in the percentage value of Ist, IInd and IIIrd Case.

Table \ref{activationGoogle_eval} shows the evaluation of 29 neurons for the remaining 20\% of the Google Image Dataset. The activation percentage for each neuron is listed for all three cases.

At this point, we can say that by our hypothesis and our verification process -- neurons get activated by the presence of concepts and concepts plays a role in deciding the output given by the network.

\begin{table}
    \centering
    \begin{tabular}{lrrr}
        \toprule
         & Case-I  & Case-II & Case-III \\
        \midrule
        neuron4     &100\%		&100\%		&100\%\\
        neuron5     &34.49\%	&37.79\%	&38.10\%\\
        neuron6     &0.5\%		&0.53\%		&0.19\%\\
        neuron7     &0.14\%		&0.19\%		&0.20\%\\
        neuron9     &100\%		&100\%		&100\%\\
        neuron11    &98.83\%	&98.83\%	&98.83\%\\
        neuron12    &94\%		&94\%		&94\%\\
        neuron13    &0.20\%		&0.22\%		&0.20\%\\
        neuron15    &100\%		&99.85\%	&100\%\\
        neuron16    &100\%		&100\%		&100\%\\
        neuron22    &35.13\%	&23.06\%	&25.62\%\\
        neuron23    &100\%		&100\%		&100\%\\
        neuron27    &99.74\%	&99.45\%	&99.77\%\\
        neuron29    &1	\%		&1\%		&1\%\\
        neuron34    &56.74\%	&56.74\%	&58.17\%\\
        neuron35    &14.29\%	&17.62\%	&9.55\%\\
        neuron36    &0.14\%		&0.14\%		&0.22\%\\
        neuron37    &0.18\%		&0.20\%		&0.13\%\\
        neuron39    &0.58\%		&0.39\%		&0.5\%\\
        neuron45    &0.12\%		&0.27\%		&0.19\%\\
        neuron52    &0.36\%		&0.19\%		&0.33\%\\
        neuron54    &0.19\%		&0.22\%		&0.12\%\\
        neuron55    &55.37\%	&46.27\%	&35.39\%\\
        neuron56    &4.82\%		&4.32\%		&2.17\%\\
        neuron58    &1.33\%		&1.33\%		&1.40\%\\
        neuron59    &0.12\%		&0.14\%		&0.22\%\\
        neuron60    &99.88\%	&100\%		&99.84\%\\
        neuron62    &100\%		&100\%		&100\%\\
        neuron63    &100\%		&100\%		&100\%\\
        \bottomrule
    \end{tabular}
    \caption{Activation percentage after doing evaluation for each neuron with Google Images for all three cases.}
    \label{activationGoogle_eval}
\end{table}

\section{Conclusion and Future Work}
\label{conclusion}
This paper is an effort toward recognizing the activation pattern of neurons in the hidden layer of CNN architecture with the presence of abstract concepts. A novel approach using ECII as an explanation generation algorithm and Wikipedia as background knowledge was shown to quantify how well a concept is recognized across the latest convolutional layer (specifically the dense layer) of a CNN. Through our verification and evaluation using Google Images, we have also reported on promising activation percentages to support our hypothesis. 

Future work will incorporate the studying of the remaining neurons and will study the effect of the different thresholds for activation of the neuron. We will need to automate the whole process of getting the human-understandable explanation for the output of the network; given the classification of the network (output of the network) as an input, it should output the activation concepts for the neurons to limit the human-intervention and explain the decision of the network efficiently.

\smallskip
\noindent\emph{Acknowledgement.} This work was supported by the U.S. Department of Commerce, National Science Foundation, under award number 2033521.

\bibliographystyle{abbrv}
\bibliography{references}

\end{document}